\documentclass[10pt,twocolumn,letterpaper]{article}

\usepackage[algorithms]{wacv}   
\usepackage{graphicx}
\usepackage{booktabs}
\usepackage{enumitem}
\usepackage{wrapfig}
\usepackage{subcaption}
\usepackage{tabularx}
\usepackage{multirow}
\usepackage{amsmath}
\usepackage{xcolor}
\usepackage{ulem}
\usepackage{balance}

\definecolor{wacvblue}{rgb}{0.21,0.49,0.74}
\usepackage[pagebackref,breaklinks,colorlinks,allcolors=wacvblue]{hyperref}

\def\wacvPaperID{332} 
\def\confName{WACV}
\def\confYear{2027}

\title{MoSTE: Mixture of Spatio-Temporal Experts for Deepfake Detection in Videos }

\author{
Charbel Yahchouchi$^{1, 2*}$\qquad
Nabyl Quignon$^{1*}$ \qquad
Noemi Roggero$^{2}$
\\[0.5em]
Laurent Saroul$^{2}$ \qquad
Antitza Dantcheva$^{1}$
\\[0.25em]
{\normalsize
$^{1}$ Inria Centre at Université Côte d'Azur \qquad
$^{2}$ Probayes
}
}

\begin{document}
\maketitle
\begingroup
\renewcommand{\thefootnote}{\fnsymbol{footnote}}
\footnotetext[1]{Equal contribution.}
\endgroup

\begin{abstract}

Generalizable deepfake video detection remains challenging because forgery artifacts vary substantially across manipulation types and often combine diverse spatial and temporal inconsistencies. Existing parameter-efficient adaptation methods typically rely on a single shared representation, rendering them prone to manipulation-specific bias and limiting transfer to unseen forgeries.
 To address this challenge, we propose \textbf{M}ixture \textbf{O}f \textbf{S}patio-\textbf{T}emporal \textbf{E}xperts (MoSTE), an adaptation framework that enables conditional specialization of forensic representations. Instead of forcing one adapter to model all forgery patterns, MoSTE decomposes adaptation into parallel spatial and temporal expert pools and dynamically routes tokens to the most relevant experts through an unsupervised token-level routing mechanism. This allows experts to naturally specialize in complementary forgery cues without requiring explicit manipulation labels or multi-stage training, while effectively transferring representations from CLIP to the video domain.
Extensive experiments across cross-dataset and cross-manipulation benchmarks demonstrate that MoSTE sets a new state of the art in average generalization while preserving strong performance across diverse manipulation families. Additional analysis of expert routing and attention behavior reveals the emergence of expert diversity, suggesting that MoSTE learns complementary spatio-temporal forensic features. Code is available at \url{https://github.com/charbelyah/MoSTE}

\end{abstract}
\section{Introduction}

Video generation \cite{wang2024lia,rombach2022high,quignon2026theval,chopin2026ai,wan2025} has reached unprecedented levels of realism, driven by remarkably rapid advancements in generative models including Stable Diffusion \cite{rombach2022high}, Sora \cite{brooks2024video}, Flux \cite{flux2024}, and WanVideo \cite{wan2025}. While such advancements allow for new creative prospects in use cases related to entertainment and marketing, they thrust upon us the danger of deepfakes - generated videos portraying individuals in scenarios that never occurred. Deepfakes pose an imminent security threat, and have already been weaponized for manipulation, disinformation, and fraud~\cite{Matich03102025, Taeihagh2025}.

\begin{figure}[t]
    \centering
    \includegraphics[width=\linewidth,height=5.5cm,keepaspectratio]{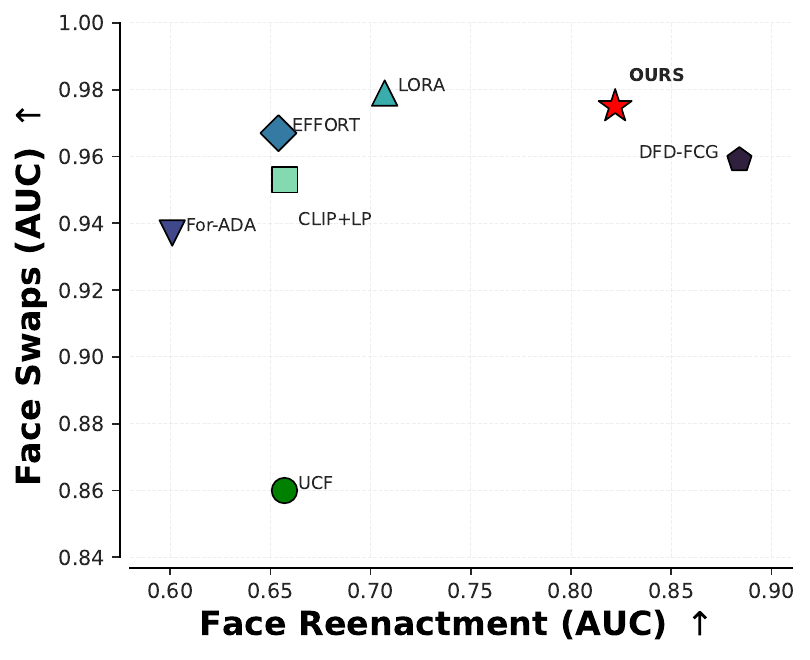}
    
    \caption{
Face Swap vs. Face Reenactment generalization on DF40 dataset. We compare MoSTE with CLIP+LP~\cite{radford2021learning}, LoRA~\cite{hu2022lora}, UCF~\cite{yan2023ucf}, DFD-FCG~\cite{han2025}, EFFORT~\cite{yan2025orthogonal}, and ForAda~\cite{cui2025forensics} . Each point reports video-level AUC averaged over the corresponding manipulation family. MoSTE, shown with a star, achieves the best balance between Face Swap and Face Reenactment performance. 
}
    \label{fig:intro}
\end{figure}

Consequently, \textit{deepfake detection} has gained high interest and impact in computer vision~\cite{qian2020thinking,11641655,atreya2026beyond,chopin2026ccdf}. Early approaches leveraged deep learning architectures aimed at discriminating real and fake images and videos, and were able to achieve impressive results in in-dataset evaluation; however, their performance often collapsed in cross-dataset settings, i.e., they \textit{lacked generalizability}. 
Therefore, more recent research aimed at improving generalization by focusing on specific forgery clues, such as \textit{frequency domain analysis} \cite{luo2021generalizing,qian2020thinking} and \textit{blending artifacts} \cite{li2020face}. Additional approaches explored the use of \textit{synthetic data} \cite{shiohara2022detecting,zhao2021learning}, in order to reduce overfitting to manipulation-specific cues.
We note that existing detectors remain challenged when facing images and videos pertaining to ``unseen'' generative models, \textit{i.e.,} out of training distribution, as they tend to overfit to specific artifact distributions of their training domains.

Recently, \textit{adapting foundation models} has become a promising direction towards more generalizable detection~\cite{han2025, cui2025forensics, yan2025orthogonal, yan2025generalizing, Cozzolino2024}, as these models demonstrate inherent zero-shot capabilities. Such parameter-efficient fine-tuning (PEFT) techniques for deepfake detection primarily focus on equipping pretrained backbones with forgery priors, leveraging the associated foundational knowledge.

However, such approaches often tend to neglect the \textit{inductive bias} inherent 
to each PEFT technique~\cite{mai2025lessons,neyshabur2014search}. 
Specifically, a model adapted using a fine-tuning strategy may become highly effective at detecting certain manipulation types, while underperforming on others. Such bias limits the robustness of detectors and hinders generalization across diverse and unseen forgery methods.
We illustrate in Fig.~\ref{fig:intro} how different adaptations yield inconsistent performance in deepfake detection, in particular for Face Swap and Reenactment. We note that Face Swap involves identity replacement, whereas Face Reenactment entails transferring facial expressions from one person to another. 
EFFORT~\cite{yan2025orthogonal} and ForAda~\cite{cui2025forensics} achieve high Area Under the Curve (AUCs) on Face Swap; however, on Face Reenactment, they show similar performance to the base CLIP model. Conversely, methods such as DFD-FCG~\cite{han2025} significantly improve results on Face Reenactment by considering temporality, whereas demonstrate limited performance on Face Swap, on challenging datasets such as DFDC \cite{dolhansky2020deepfake}.

We note that it is imperative to  
consider the \textit{bias limits}, in order to ensure generalizability. 
Therefore, we deviate from a  
single representation by enforcing independent learning of different forgery traces and modeling thereof.  
Specifically, we decompose adaptation into specialized \textit{spatial and temporal experts} and allow tokens to \textit{dynamically select} the most relevant experts during training. This transforms adaptation from a fixed global transformation into an input-dependent process.

We propose a Mixture of Spatio-Temporal Experts (MoSTE), where we insert an adapter across transformer layers of a frozen CLIP-ViT-L/14 backbone. MoSTE disentangles adaptation into spatial and temporal expert pools, while token-level routing dynamically assigns tokens to the most relevant experts. This enables specialization across both artifact nature and artifact diversity without explicit manipulation labels. This approach effectively resolves the imbalance in manipulation sensitivity and establishes a highly generalizable detection framework. Our contributions include the following. 

\vspace{0.2cm}
\begin{itemize}[left=5pt,noitemsep,nolistsep]
  \item We introduce MoSTE, a \textit{novel single-stage PEFT framework for video deepfake detection} that performs \textit{spatio-temporal adaptation} through \textit{token-level expert routing}. Unlike previous approaches that learn a single shared representation per manipulation type, MoSTE allows unsupervised specialization across diverse forgery traces without explicit guidance.

 \item We conduct extensive experiments, comparing MoSTE to recent SOTA on datasets such as Celeb-DF-v3 \cite{li2025celeb} and DF40 \cite{yan2024df40} under a set of protocols. Our results suggest that \textit{MoSTE is endowed with generalizability}, as opposed to SOTA models. 
    \item We evaluate MoSTE  
    entailing large-scale training using the DF40 Face Swap and Face Reenactment subsets, in order to assess the scalability of our proposed adaptation architecture and its generalization capability.
\end{itemize}

\section{Related work}

\subsection{Generalizable Deepfake Detection}
Limited generalization capacity of deepfake detection stems from the tendency of models to capture manipulation-specific cues rather than intrinsic forensic traces. Towards mitigating this challenge, a prominent research direction focused on learning universal forgery artifacts, including frequency-domain inconsistencies \cite{luo2021generalizing,liu2021spatial,qian2020thinking}, blending artifacts \cite{li2020face,cui2025forensics,nguyen2024laa}, and facial attribute anomalies \cite{haliassos2021lips,guo2022eyes}. Another widely explored approach leveraged blending data \cite{li2020face,shiohara2022detecting,zhao2021learning} by simulating common forgery inconsistencies, aimed at encouraging models to learn generalized forgery features. Subsequent studies \cite{cheng2024can,lin2024fake} have proposed training strategies that combine handcrafted pseudo-fake samples and deepfake data from FaceForensics++ (FF++) dataset.  
Under the same blending-data paradigm, several works \cite{nguyen2025vulnerability,cai2025deepshield,yan2025generalizing} extended pseudo-fake generation to the video domain by simulating common temporal artifacts, like temporal landmark drifting. 

\textit{Tackling generalizability}, research works \cite{gu2021spatiotemporal,Kim_2025_ICCV,zhang2024learning} advanced deepfake detection into the spatio-temporal domain by explicitly modeling temporal dynamics to capture artifacts such as temporal inconsistency, and unnatural motion. These methods employ diverse strategies, including 3D convolutional backbones \cite{zheng2021exploring}, temporal encoding schemes \cite{xu2023tall}, spatial–temporal optimization \cite{wang2023altfreezing}, and latent style-flow cues \cite{choi2024exploiting}.

Deviating from the aforementioned approaches, we assume that each manipulation category exhibits distinct forgery artifacts, and therefore a single adaptation strategy is insufficient. To address this, we propose a novel adapter architecture capable of dynamically adapting to the diverse landscape of forensic traces.

\subsection{Foundation Model Adaptation for Deepfake Detection}
Foundation vision-language models such as CLIP~\cite{radford2021learning} have strong transfer ability, motivating recent CLIP-based deepfake detectors~\cite{cui2025forensics,yan2025orthogonal,yan2025generalizing,han2025,yahchouchi2026generalizable}. These methods preserve pretrained representations while learning forensic cues~\cite{yan2025orthogonal}: ForensicsAdapter~\cite{cui2025forensics} learns blending-aware attention biases, Yan et al.~\cite{yan2025generalizing} add spatial-temporal modules using synthetic videos, DFD-FCG~\cite{han2025} uses a side decoder for intermediate spatial-temporal cues, and EFFORT~\cite{yan2025orthogonal} adapts LoRA~\cite{hu2022lora} while retaining principal components. However, these PEFT strategies remain fixed across manipulation types and may introduce their own \textit{inductive bias}~\cite{mai2025lessons}, limiting generalization to out-of-training manipulations. MoSTE instead routes tokens through specialized spatial and temporal expert pools.
\subsection{Mixture of Experts in Deepfake Detection}
The Mixture of Experts (MoE) paradigm has recently emerged in multimedia forensics as an effective mechanism to improve cross-domain generalization. MoE-FFD~\cite{Moeffd} integrates LoRA and Adapter layers as distinct experts within Vision Transformers (ViT) to extract complementary global and local forgery cues. DevFD~\cite{zhang2025devfd} extends this to continual learning by adding orthogonal LoRA experts for emerging manipulation types. Forensic-MoE~\cite{fang2025forensic} broadens the scope to universal synthetic image detection by assigning experts explicitly to GAN and Diffusion generative domains.
Despite their promise, these approaches share two fundamental limitations. First, they operate at the frame level~\cite{Moeffd, zhang2025devfd, fang2025forensic}, entirely discarding temporal artifacts. Second, they rely on explicit domain supervision or multi-stage training to guide expert specialization~\cite{zhang2025devfd, fang2025forensic}. In contrast, MoSTE addresses both limitations through a single-stage spatio-temporal framework with an unsupervised routing mechanism that autonomously discovers spatial and temporal forgery traces without requiring domain labels or sequential pre-training.

\section{Methodology}

\subsection{Overview} 
\begin{figure*}[t]
\centering
\includegraphics[width=0.9\textwidth]{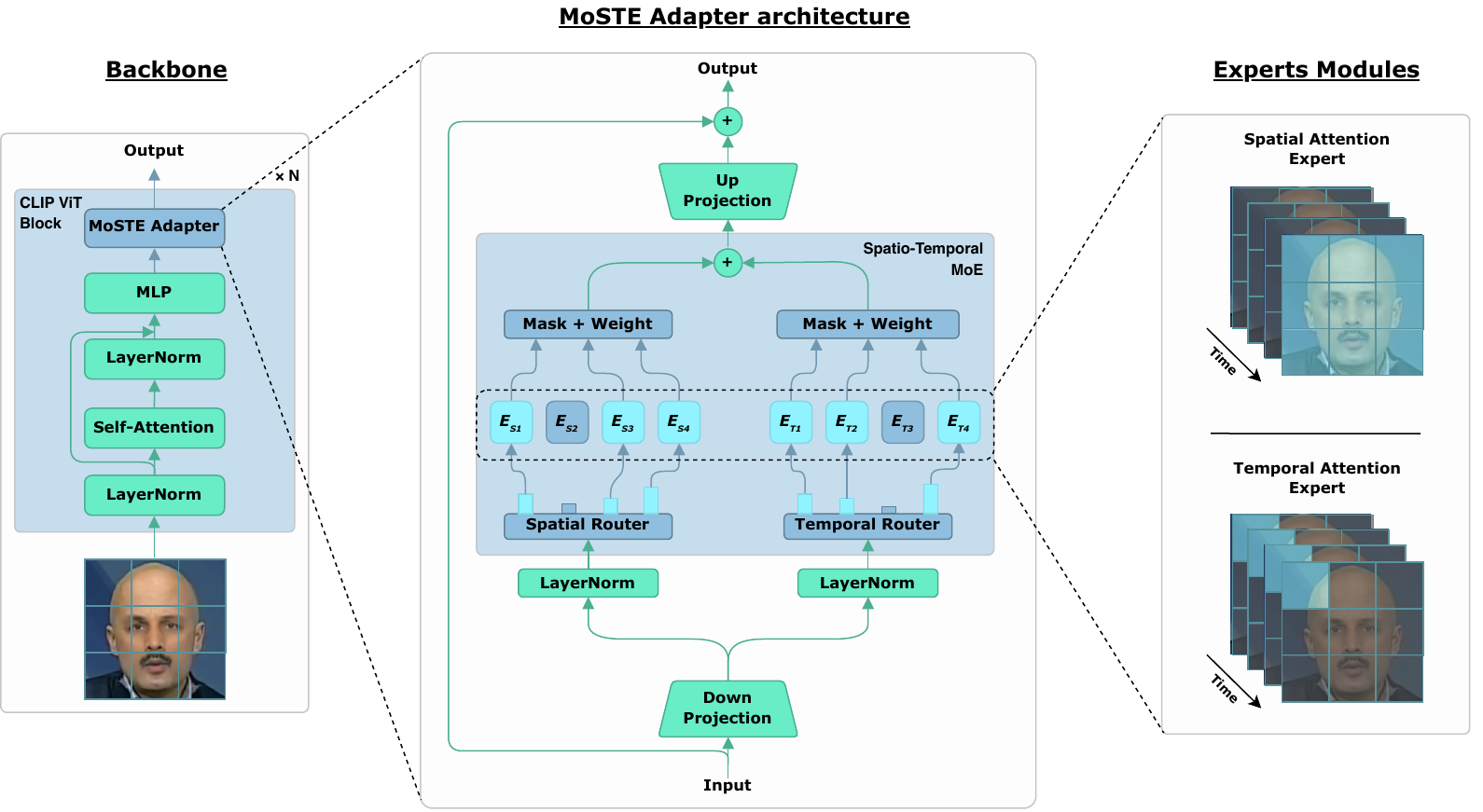}
\caption{\textbf{Overview of the MoSTE architecture.} MoSTE adapters are inserted into a frozen CLIP-ViT backbone and operate in the adapter bottleneck space. Tokens are routed to independent spatial and temporal expert pools. The selected expert outputs are masked, weighted, fused, and projected back through a residual adapter update.  }
\label{fig:architecture_comparison}
\end{figure*}

In this work, we introduce MoSTE, an adapter to transfer frozen CLIP-ViT image representations to the video deepfake detection domain. The key challenge is that forgery evidence is diverse: some artifacts are spatial and appear within individual frames, while others are temporal and only emerge across frames. Moreover, different manipulation families introduce different traces, e.g., face swaps often produce blending artifacts, whereas reenactment methods may create motion or landmark inconsistencies. A single shared adapter projection has to encode all these cues with the same parameters, which can limit specialization. MoSTE addresses this diversity by replacing the standard adapter bottleneck with a mixture of spatio-temporal experts. Each adapter first projects tokens to a compact bottleneck representation, then routes them to two expert pools: spatial experts process patches within each frame, while temporal experts process the same patch locations across frames. A token-level router selects and weights the most relevant experts, allowing different tokens to rely on different forensic cues without requiring manipulation labels. Fig.~\ref{fig:architecture_comparison} illustrates the overall adapter framework. Section~\ref{sec:MOE} details the expert design, and Section~\ref{sec:routig} describes routing and load balancing.

\subsection{Spatio-Temporal Expert Design} \label{sec:MOE} MoSTE builds on the observation that artifacts are both spatially and temporally different. Instead of using a fixed dual-pathway adapter~\cite{park2023dual,pei2025d2st}, we introduce conditional spatial and temporal expert pools with token-level routing. Each adapter maps the input $X_{in}$ to a bottleneck representation $X_{mid}$ through a shared down-projection and GELU activation, then processes $X_{mid}$ with two independent expert pools. \paragraph{Spatial and temporal experts.} We define two expert pools, $\mathcal{E}_{B}=\{E^{B}_{e}\}_{e=1}^{N}$ with $B \in \{S,T\}$, where $S$ and $T$ denote the spatial and temporal branches, respectively. Each expert is implemented as self-attention over a branch-specific token arrangement: \begin{equation} E^{B}_{e}(X_{mid}) = \mathrm{Attn}_{B,e}\big(\mathrm{Reshape}_{B}(X_{mid})\big), \quad B \in \{S,T\}. \end{equation} Here, $\mathrm{Reshape}_{S}$ groups patches within each frame to capture intra-frame artifacts, while $\mathrm{Reshape}_{T}$ groups the same patch location across frames to capture inter-frame inconsistencies.

\begin{figure}[t]
    \centering
    \begin{subfigure}{0.5\textwidth}
        \centering
        \includegraphics[width=\linewidth]{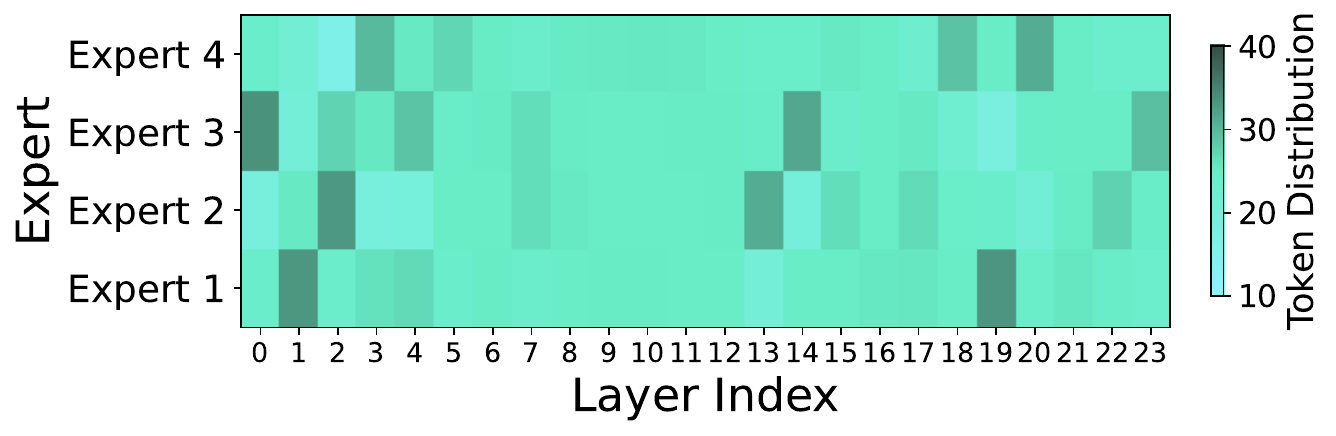}
        \caption{Spatial distribution}
        \label{fig:routing_spatial}
    \end{subfigure}
    \hfill 
    \begin{subfigure}{0.49\textwidth}
        \centering
        \includegraphics[width=\linewidth]{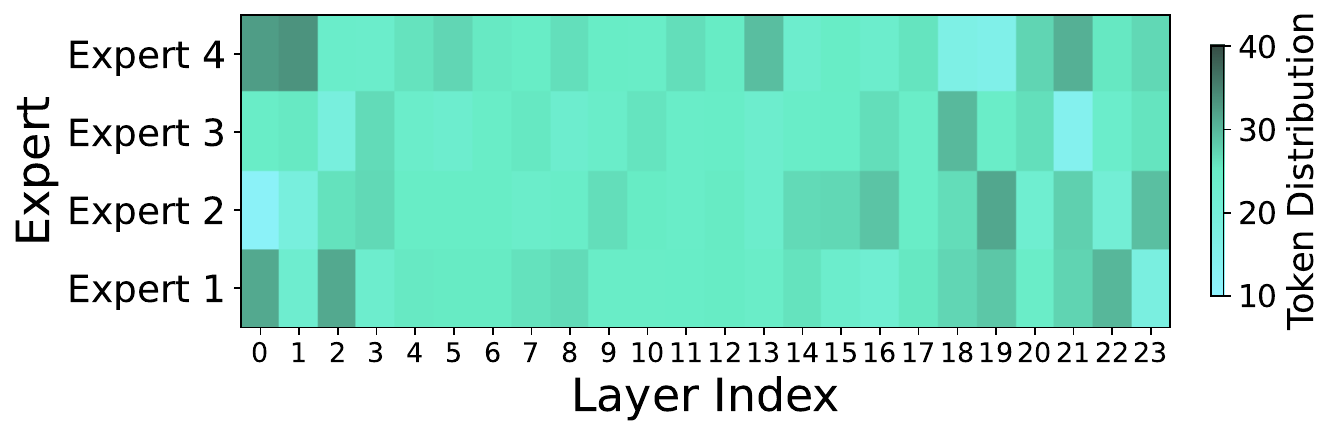}
        \caption{Temporal distribution}
        \label{fig:routing_temporal}
    \end{subfigure}
    
    \caption{\textbf{Expert routing load}. Each cell shows the percentage of tokens routed to each of the four experts (E1–E4), with separate heatmaps for spatial and temporal attention. A perfectly balanced distribution corresponds to 25\% per expert. The middle layers (4-12) exhibit near-uniform routing, while the early (0-2) and late (18-23) layers show stronger specialization.}
    \label{fig:moe_routing_analysis}
\end{figure}

\subsection{Expert Routing and Specialization}
\label{sec:routig}

\paragraph{Token-level routing.}
The goal of routing is \textit{specialization}: each expert should consistently handle a 
specific type of forensic trace. For each token $i$ in $X_{mid}$, a linear gating network produces routing logits $r_{i,e}$ 
over the $E$ available experts. The Top-$k$ experts are selected per 
token, and routing weights are computed via softmax restricted to those 
selected experts:
\begin{equation}
    w_{i, e} = 
    \begin{cases} 
        \dfrac{\exp(r_{i, e})}{\sum_{j \in \mathcal{T}_i} \exp(r_{i, j})} 
        & \text{if } e \in \mathcal{T}_i \\[8pt]
        0 & \text{otherwise}
    \end{cases},
    \label{eq:routing_weights}
\end{equation}
where $\mathcal{T}_i$ is the set of Top-$k$ selected experts for token 
$i$. Each active expert processes the full sequence, but its contribution 
is gated by these weights. Tokens not assigned to a given expert 
receive a zero weight, effectively masking its output. The aggregated 
spatial branch output is:
\begin{equation}
    Y_{S} = \sum_{e=1}^{N} w_{i, e} \odot \mathcal{E}_{S, e}(X_{mid}),
    \label{eq:spatial_output}
\end{equation}

and $Y_{T}$ is computed identically for the temporal branch.
The outputs of both branches are additively fused with $X_{mid}$, passed 
through a GELU activation, and projected back to the original dimension 
via a shared up-projection $W_{up}$, with a residual connection 
preserving the original feature stream:
\begin{equation}
    X_{out} = X_{in} + 
        \text{GELU}\!\left(X_{mid} + Y_{S} + Y_{T}\right) W_{up} 
    \label{eq:adapter_output}
\end{equation}
\paragraph{Load balancing without auxiliary losses.}
A common failure mode in MoE models is \textit{routing collapse}, where most tokens are repeatedly assigned to a small subset of experts. This reduces expert diversity and limits the benefit of using an MoE architecture. Standard approaches often prevent this behavior with auxiliary load-balancing losses, but these losses introduce an additional optimization objective that may interfere with the main deepfake detection loss.

To avoid this, we follow the auxiliary-loss-free strategy of DeepSeek-V3~\cite{deepseek2025}. Each expert $e$ is assigned an adaptive routing bias $b_e$, which is used only when selecting the Top-$k$ experts for each token:
\begin{equation}
    \mathcal{T}_i = \text{Top-}k_e(r_{i,e} + b_e),
    \label{eq:biased_topk}
\end{equation}
where $r_{i,e}$ is the routing logit between token $i$ and expert $e$, and $\mathcal{T}_i$ denotes the selected expert set.

Importantly, the bias affects only the expert selection step. Once the experts are selected, the routing weights are still computed from the original logits $r_{i,e}$, as in Eq.~\ref{eq:routing_weights}. This preserves the main learning signal while allowing the router to correct imbalanced expert usage.

After each training step, we update the bias according to the observed expert load. Experts that receive too many tokens have their bias decreased, making them less likely to be selected in subsequent steps. Conversely, underused experts have their bias increased, encouraging the router to assign them more tokens. This simple feedback mechanism keeps expert utilization balanced without introducing any auxiliary loss. As shown in Fig.~\ref{fig:moe_routing_analysis}, it maintains well-distributed routing across experts while allowing the model to optimize only the primary detection objective.

\section{Experiments}
\label{sec:experiments}
We here elaborate on implementation details and evaluation settings,
and proceed to evaluate MoSTE under two deepfake detection protocols, aimed at assessing generalization across datasets and manipulation types. We then conduct an ablation study, in order to quantify the contribution of each MoSTE-component.

\subsection{Implementation Details}
\label{sec:impl_details}

We use CLIP-ViT-L/14~\cite{radford2021learning} as a frozen backbone and insert MoSTE adapters into all 24 transformer blocks, optimizing only the adapters, classification head, and learnable temporal embeddings. Each adapter has bottleneck dimension $d_{\text{adapter}}=128$ and separate spatial/temporal expert pools with $N=4$ experts each, Top-$k$ routing with $k=3$, and auxiliary-loss-free bias balancing with update speed $\gamma=10^{-4}$. We train for 10 epochs on 4 NVIDIA H100 GPUs using Adam with learning rate $4\times10^{-4}$, $\beta_1=0.9$, $\beta_2=0.999$, weight decay $5\times10^{-4}$, and cosine annealing to $10^{-5}$ with $t_{\max}=10$. From each video, we sample 32 frames and feed 8-frame clips with horizontal flip, random rotation, Gaussian blur, and brightness/contrast perturbations.

\subsection{Evaluation Settings}

We conduct training on the FF++ dataset~\cite{rossler2019faceforensics++} under the C23 compression setting. The training split comprises 1,000 real and 4,000 manipulated videos.

To comprehensively evaluate the generalization capability of the proposed approach, we adopt two complementary evaluation protocols, namely Protocol-1 (cross-dataset evaluation) and Protocol-2 (cross-manipulation evaluation within the FF++ domain) that we proceed to elaborate on.
\begin{table*}[t]
\centering
\caption{Cross-dataset evaluation results under Protocol-1. All methods are trained on FaceForensics++ (FF++) C23 and evaluated across different datasets. Baseline results are taken from \cite{yan2025orthogonal} and reported in terms of video-level AUC (\%). $^\dagger$ indicates inference from official checkpoints. The best result for each dataset is highlighted in bold, and the second-best result is underlined.}
\label{tab1}
\begin{tabularx}{\textwidth}{l|*{6}{>{\centering\arraybackslash}X}}
\toprule
Detector  & CDF-v2 & CDF-v3$^\dagger$ & DFD & DFDC & DFDCP & Avg \\
\midrule
SBI \cite{shiohara2022detecting}      
& 93.2 & -- & 82.7  & 72.4 & -- & 82.8 \\

UCF \cite{yan2023ucf}      
& 83.7 & 83.6 & 86.7 & 74.2 & 77.0 & 81.0 \\

LSDA \cite{yan2024transcending}     
& 87.5 & -- & 88.1  & 70.1 & 81.2 & 81.7 \\

ProDet  \cite{cheng2024can}    
& 92.6 & -- & 90.1  & 70.7 & 82.8 & 84.1 \\

CDFA \cite{lin2024fake}  
& 93.8 & -- & 95.4  & 83.0 & 88.1 & 90.1 \\

UDD \cite{fu2025exploring}      
& 93.1 & -- & 95.5  & 81.2 & 88.1 & 89.5 \\

FakeSTormer \cite{nguyen2025vulnerability} & 92.4 & -- & 96.5  & 74.6 & 90.0 & 88.4 \\
ForAda \cite{cui2025forensics}    
& \textbf{95.7} & 84.5 & \underline{97.2}  & \textbf{87.2} & \textbf{92.9} & 91.5 \\

DFD-FCG \cite{han2025}     
& 95.0 & \textbf{93.8} & --  & 81.8 & -- & 90.2 \\
GenD \cite{yermakov2026deepfake}    
& 95.4 & 90.1 & 96.4  & 86.4 & 90.0  & \underline{91.7}  \\

EFFORT $^\dagger$ \cite{yan2025orthogonal}    
& 93.2 & 87.2 & 96.5  & 84.3 & 85.6 & 89.4 \\

\midrule

Ours (MoSTE)                   
& \underline{95.5} & \underline{91.3} & \textbf{97.5} & \underline{86.8} & \underline{90.7} & \textbf{92.4} \\

\bottomrule
\end{tabularx}
\end{table*}

\noindent\textbf{Protocol-1. Cross-Dataset Evaluation.}
In this setting, models trained on FF++ are evaluated on large-scale deepfake benchmarks without further fine-tuning. Specifically, we consider Celeb-DF v2 (CDF-v2)~\cite{li2020celeb}, Celeb-DF-v3 (CDF-v3)~\cite{li2025celeb}, DeepFakeDetection (DFD)~\cite{dufour2019contributing}, the DeepFake Detection Challenge (DFDC) dataset~\cite{dolhansky2020deepfake}, and its preview version (DFDCP)~\cite{dolhansky2019deepfake}. This protocol measures robustness to domain shifts in data distribution, compression artifacts, and manipulation pipelines.

\noindent\textbf{Protocol-2. Cross-Manipulation Evaluation.}
As proposed by Yan \textit{et al.}~\cite{yan2025orthogonal}, we further assess generalization to unseen manipulation methods using the recently introduced DF40 dataset~\cite{yan2024df40}. DF40 is constructed from real samples in CDF and includes deepfakes generated with 40 additional manipulation techniques, including face swaps and reenactment. Protocol-2 evaluates the model's ability to generalize to a more diverse set of forgery methods.

\subsection{Generalization Evaluation}
\subsubsection{Cross-Dataset Evaluation}

\textit{W.r.t.} Protocol-1, all methods are trained on FF++ and directly evaluated on unseen target datasets to assess cross-dataset generalization. The results are reported in Table~\ref{tab1} in terms of video-level AUC (\%). Our results suggest that MoSTE achieves superior average performance across datasets, indicating strong and consistent generalization ability. In particular, MoSTE attains the top performance on DFD and ranks second on CDF-v3 and DFDC, while remaining highly competitive on the other benchmarks. Although ForAda achieves the best results on CDF-v2, DFDC, and DFDCP, our approach demonstrates more balanced performance across datasets.

Compared with existing SOTA, MoSTE excels on single benchmarks, as well as maintains robust performance under different distribution shifts, i.e., ability to generalize across datasets.

\subsubsection{Cross-Manipulation Evaluation}
\textit{W.r.t.} Protocol-2, where models are trained on FF++ and evaluated on unseen DF40 manipulations, Table~\ref{tab:cross_method} directly validates our central claim by evaluating both Face Swap and Face Reenactment attacks. MoSTE achieves the best overall performance and maintains a consistent high AUC across both manipulation families.

Unlike Fig.~\ref{fig:intro}, which averages over all DF40 methods per family, Table~\ref{tab:cross_method} reports per-method results on a representative subset covering distinct generation mechanisms (e.g., latent-space animation in LIA, audio-driven lip synchronization in Wav2Lip, warping-based motion transfer in TPSM). MoSTE performs strongly on Face Swap (BleFace, InSwap, SimSwap, UniFace) while generalizing well to Face Reenactment (Wav2Lip, LIA, TPSM), whereas recent PEFT-based detectors are less balanced across the two families. This indicates that spatio-temporal expert routing reduces manipulation-specific bias and yields more transferable forensic representations.

\begin{table*}[t]
\centering
\caption{Cross-manipulation evaluation under Protocol-2. Models are trained on FF++ and tested on CDF-domain face swap and reenactment manipulations. All methods use official checkpoints and setup settings provided by authors. Metric: video-level AUC (\%). Best results are bolded and second-best underlined.}
\label{tab:cross_method}
\begin{tabularx}{\textwidth}{l|*{9}{>{\centering\arraybackslash}X}}
\toprule
& \multicolumn{5}{c}{Face Swap} & \multicolumn{4}{c}{Face Reenactment} \\
\cmidrule(lr){2-6} \cmidrule(lr){7-10}
Method & BleFace & Inswap & Simswap & UniFace & Avg. & Wav2Lip & LIA & TPSM & Avg. \\
\midrule

UCF  \cite{yan2023ucf}     
& 66.3 & 70.5 & 47.3 & 66.7 & 62.7 & 71.0 & 87.2 & 57.0 & 71.7 \\

ProDet \cite{cheng2024can}  
& 83.1 & 75.1 & 64.4 & 78.4 & 75.3 & 71.5 & 75.1 & 46.7 & 64.4 \\

DFD-FCG \cite{han2025}  
& \textbf{95.4} & \underline{93.7} & \underline{92.7} & \textbf{97.9} & \underline{94.9} & 57.4 & \textbf{98.5} & \textbf{89.5} & \underline{81.8} \\

GenD \cite{yermakov2026deepfake}    
& 92.8 & 89.1 & 90.1 & 97.0 & 92.3 & \textbf{78.9} & 88.8 & 70.8 & 79.5 \\

EFFORT \cite{yan2025orthogonal} 
& 89.7 & 87.8 & 86.9 & 92.8 & 89.3 & 64.5 & 79.6 & 62.2 & 68.8 \\

ForAda \cite{cui2025forensics} 
& 82.8 & 89.1 & 75.9 & 86.0 & 83.4 & 57.8 & 79.5 & 62.7 & 66.7 \\

\midrule
Ours (MoSTE) 
& \underline{93.7} & \textbf{96.3} & \textbf{95.7} & \underline{97.7} & \textbf{95.9} & \underline{78.4} & \underline{97.4} & \underline{82.4} & \textbf{86.1} \\
\bottomrule
\end{tabularx}
\vspace{-7pt}
\end{table*}
\subsubsection{Robustness Evaluation}
\begin{figure*}[t]
\centering
\includegraphics[width=\textwidth]{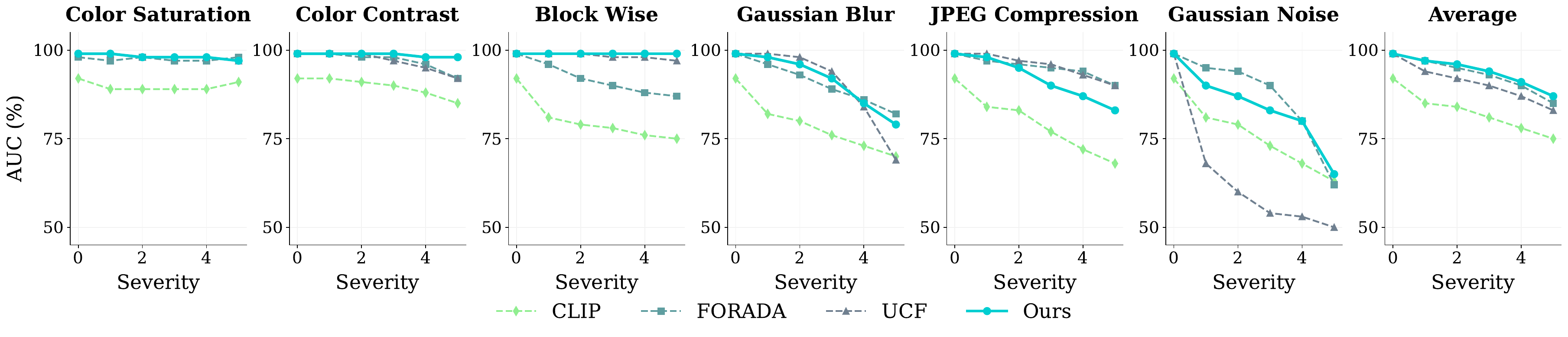}
\caption{\textbf{Robustness analysis on FF++.} We evaluate the video-level AUC under six common perturbations such as Gaussian Blur and JPEG compression across multiple severity levels. MoSTE exhibits lower performance degradation compared to SOTA. }
\label{fig:robustness_results}
\end{figure*}

We evaluate the robustness of our proposed framework against common post-processing perturbations. Following the experimental protocol in \cite{jiang2020deeperforensics}, we subject FF++ test data to six categories of distortions across multiple severity levels, \textit{viz.} color saturation, color contrast, block-wise distortions, Gaussian blur, JPEG compression, and Gaussian noise. We compare MoSTE with recent SOTA approaches, including ForAda \cite{cui2025forensics}, UCF \cite{yan2023ucf}, and the CLIP-based baseline \cite{radford2021learning}. As summarized in Fig. \ref{fig:robustness_results}, our method consistently outperforms competing approaches in mean performance. Specifically, this disentangled spatio-temporal experts setup maintains forensic fidelity despite compression and noise, demonstrating robustness to common real-world degradations.

\begin{table}[t]
    \centering
    \setlength{\tabcolsep}{5pt}
    
    \caption{\textbf{Main components ablation.} Impact of MoE, ``S'' for Spatial, and ``T'' for Temporal components. The metric is video-level AUC (\%).}
    \resizebox{\columnwidth}{!}{%
    \begin{tabular}{w{c}{0.5cm}w{c}{0.5cm}w{c}{0.5cm}|c|c|c|c|c|c}
        \toprule
        MoE & S & T & CDF-v2 & DFDCP & DFDC   &DFD & Avg. \\
        \midrule
        $\times$   & \checkmark & \checkmark & \textbf{95.5} & 88.0 & 86.5 & 97.1 & 91.8\\
        \checkmark & $\times$   & \checkmark   & 91.6 &88.1 &80.9 &94.7 &88.8\\
        \checkmark & \checkmark & $\times$   & 91.3& 85.6&80.6 &  97.4&88.7\\
        \checkmark & \checkmark & \checkmark & \textbf{95.5} & \textbf{90.7}& \textbf{86.8}& \textbf{97.5}&\textbf{92.6}\\
        \bottomrule
    \end{tabular}
    }
    \label{tab:main_components_abla}
\end{table}


\subsection{Ablation Study}  
\subsubsection{Ablation on Adapter Components}
We isolate the core components of the MoSTE architecture to validate our design choices, see 
Table~\ref{tab:main_components_abla}. Replacing the MoE routing mechanism with an unweighted ensemble of 4 experts leads to a notable degradation of generalizability, particularly on DFDCP dataset with a $-2.7\%$. This indicates that the concurrent activation of all learned features introduces noise, degrading model performance. Without the specialized focus of the routers, the final representation of each token becomes less informative, preventing the model from isolating the specific spatial or temporal traces it has learned to identify. Furthermore, disabling either the spatial or temporal branch severely degrades performance, demonstrating that both branches are mandatory.

\subsubsection{Ablation on Experts}
\begin{table}[t]
    \centering
    \setlength{\tabcolsep}{5pt}
    
    \caption{\textbf{Ablation Study on Expert Configuration.} Impact of the total number of experts ($N$) and the Top-$k$ routing selection on video-level AUC.}
    
    \begin{tabular}{c|c|c|c|c|c}
        \toprule
        Experts  & Top-$k$& CDF-v2 & DFDCP & DFDC & Avg. \\
        \midrule
        \multirow{1}{*}{$N=1$} & 1 & 92.8 & 89.8 & 85.2 & 89.3  \\
        \midrule
        \multirow{2}{*}{$N=2$} & 1 & 95.2 & 90.8 & 85.2 & 90.4 \\
                               & 2 & 95.1 & 91.0 & 86.0 & 90.7 \\
        \midrule
        \multirow{3}{*}{$N=4$} & 2 & 96.2 & 86.4 & 82.7 & 88.4 \\
                               & 3 & 95.5 & 90.7 & 86.8 & \underline{91.0} \\ 
                               & 4 & 95.8 & 88.7 & 85.5 & 90.0 \\ 
                               
        \midrule
        \multirow{2}{*}{$N=8$} & 4 & 93.6 & 89.0 & 84.1 & 88.9 \\ 
                               & 6 & 94.4 & 90.3 & 86.6 & 90.4 \\ 
        \bottomrule
    \end{tabular}
    \label{tab:expert_abla}
\end{table}

Table~\ref{tab:expert_abla} shows the effect of the number of experts $N$ and the Top-$k$ routing threshold. The experiment $N=1$ serves as a baseline; it refers to a simple dense spatio-temporal adapter without any router.
Introducing multiple experts ($N \geq 2$) consistently improves the average AUC over the single path spatio-temporal baseline, confirming the benefit of specialization. Performance increases up to $N=4$, while further scaling to $N=8$ yields no additional gains and slightly reduces stability across datasets, indicating diminishing returns from excessive capacity.

Overall, $N=4$ with Top-$k=3$ achieves the best trade-off between performance and parameter efficiency, delivering the highest average AUC while keeping the parameter increase below 7\% (19M params).
\subsubsection{One Router Ablation}
\begin{table}[t]
    \centering
    \setlength{\tabcolsep}{5pt}
    
    \caption{\textbf{One router ablation.} Comparison of expert selection methods with only one router. The metric is video-level AUC.}
    \resizebox{\columnwidth}{!}{%
    \begin{tabular}{l|c|c|c|c|c|c|}
        \toprule
        Selection  & CDF-v2 & CDF-v3 & FSh & DFD & DFDC & Avg. \\
        \midrule
        PS  &94.7 &88.1 & 87.2& 96.9& 84.7& 90.3\\
        FS & 92.3& 87.4 & 88.9 &96.7 & 84.1&89.9 \\
        Ours & \textbf{95.5}& \textbf{91.3} &\textbf{90.1} & \textbf{97.5}& \textbf{86.8}& \textbf{92.2}\\
        \bottomrule
    \end{tabular}
    }
    \label{fig:one_router_abla}
\end{table}
A key design element of MoSTE is the use of independent routers for the spatial and temporal expert pools. We evaluate this architecture against two ``Unified Routing" baselines to justify the decoupling of forensic traces. In the \textit{Pair Selection} (PS) configuration, a single router dispatches tokens to a fixed pair consisting of one spatial and one temporal expert. Conversely, the \textit{Free Selection} (FS) setting employs a single router that is free to route tokens to any available expert, regardless of whether that expert specializes in spatial or temporal features.
Results are reported in Table~\ref{fig:one_router_abla}.

\subsection{Qualitative Analysis of Expert Specialization}
\subsubsection{Spatial Expert}
\begin{figure}[t]
    \centering
    \includegraphics[width=\linewidth]{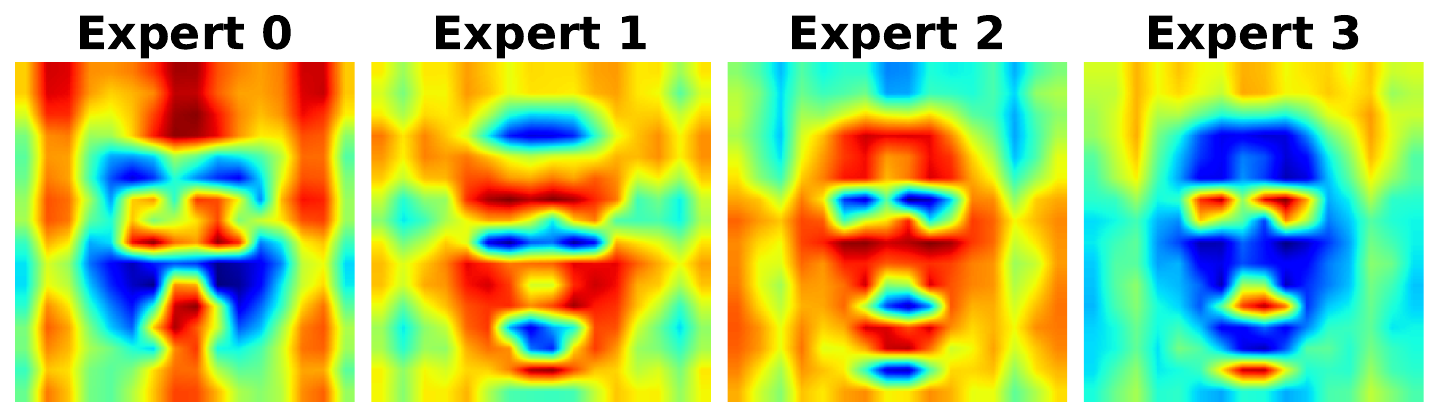}
    \vspace{1mm}
    \includegraphics[width=\linewidth]{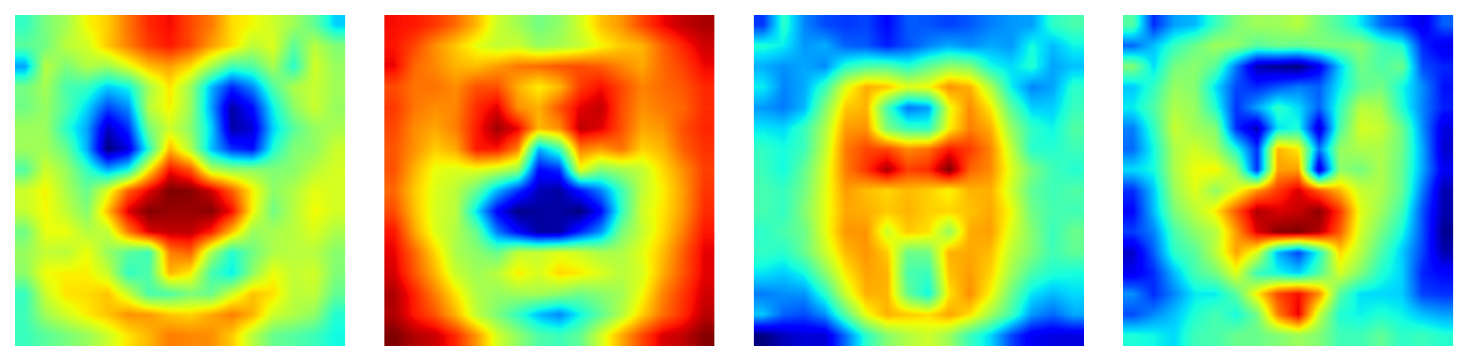}
    \vspace{1mm}
    \includegraphics[width=\linewidth]{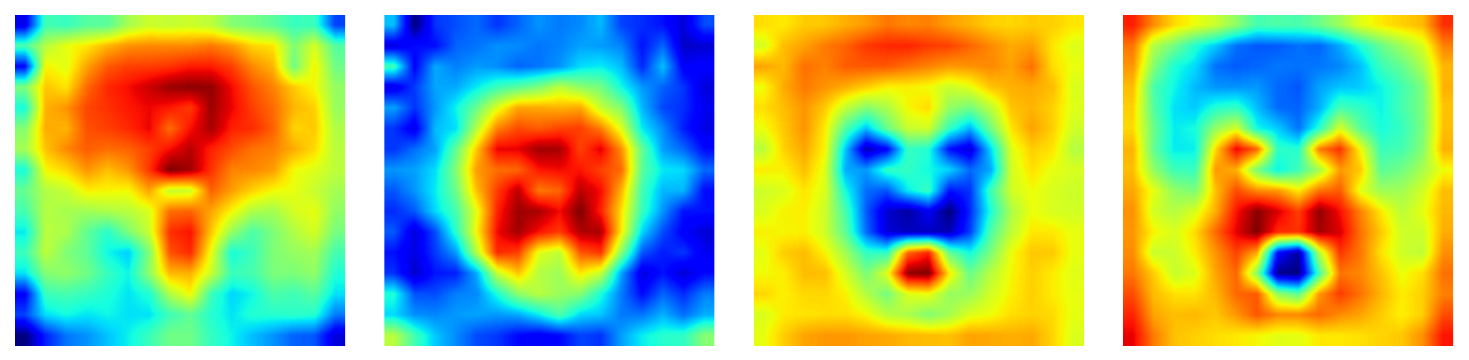}
    
    \caption{\textbf{Average Attention Maps on CDF-v2.} Visualization of the average spatial attention maps from our MoSTE adapter across three network depths (ordered from top to bottom: layers 1, 14, and 23). The heatmaps illustrate the emergent routing behavior of four distinct experts (E0–E3). }
    \label{fig:expert_maps}
\end{figure}
In Fig.~\ref{fig:expert_maps}, we visualize the averaged attention maps of four spatial experts from our MoSTE framework on CDF-v2 test samples. We select attention maps from three different backbone depths. The results show that, at each depth, the experts focus on distinct facial regions, such as the forehead, cheeks, and eyebrows, without explicit supervision. These visualizations offer indicative evidence of expert diversity, suggesting that MoSTE captures complementary cues across heterogeneous forgery features in videos.
\subsubsection{Temporal Expert Complementarity}
We use linear CKA \cite{kornblith2019similarity} to measure whether temporal experts learn redundant or complementary representations. Expert outputs are extracted before spatial fusion, up-projection, and the global residual connection, then weighted by their temporal router assignments while keeping the same ordered spatio-temporal tokens across experts.

\begin{figure}
    \centering
    \includegraphics[width=\linewidth,height=5.5cm,keepaspectratio]{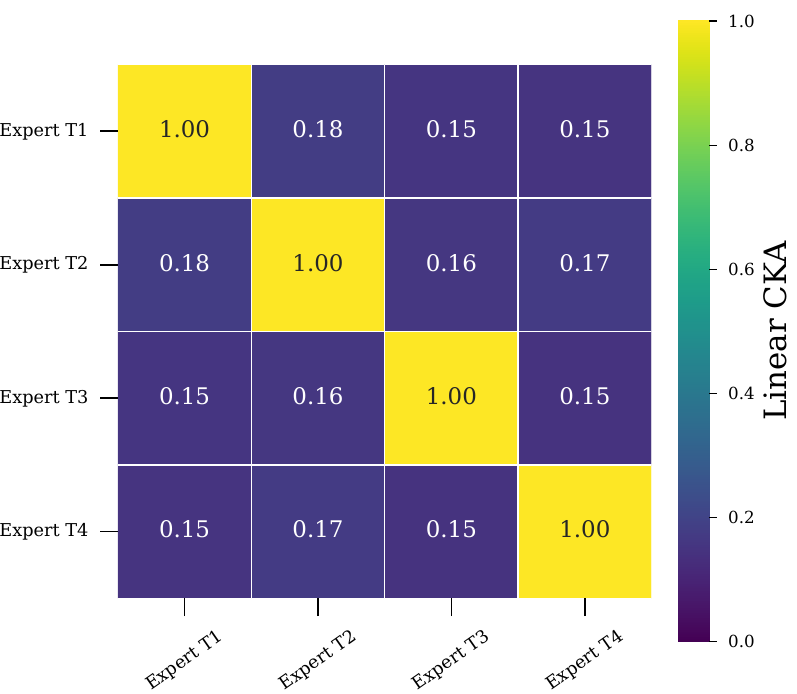}
    \caption{Temporal expert similarity measured with linear CKA. Metric is averaged over all blocks of backbone and 1000 samples from CDF-v2.}
    \label{fig:temporal_cka}
\end{figure}

As shown in Fig.~\ref{fig:temporal_cka}, off-diagonal CKA values remain low ($0.15$--$0.18$), while diagonal values are $1.0$ by construction. This suggests that temporal experts avoid representational collapse and learn complementary inter-frame forensic cues.
\subsection{Capacity-Controlled Adapter Ablation}
A natural question is whether MoSTE's gains stem from expert routing or simply from increased adapter capacity. To isolate this factor, we compare MoSTE against dense dual-pathway spatio-temporal adapters with enlarged bottleneck dimensions, chosen to match or exceed the trainable parameter count of MoSTE. As shown in Table~\ref{tab:adapter_scaling}, increasing the capacity of a dense adapter does not consistently improve out-of-distribution generalization. In particular, the 75M-parameter baseline underperforms the smaller 25M baseline on average, suggesting that additional capacity alone may amplify overfitting to the source distribution. In contrast, MoSTE achieves the best average AUC with fewer parameters than the largest dense baseline, indicating that the improvement comes from conditional spatio-temporal expert routing rather than parameter count alone.

\begin{table}[t]
\centering
\caption{Capacity-controlled comparison with dense dual-pathway adapters.}
\label{tab:adapter_scaling}
\setlength{\tabcolsep}{5pt}
\renewcommand{\arraystretch}{1.08}
\resizebox{\columnwidth}{!}{%
\begin{tabular}{@{}lccccc@{}}
\toprule
Method & Params & CDF-v2 & DFDCP & CDF-v3 & Avg. \\
\midrule
Dense ST adapter, dim=512 & 75M & 93.6 & 91.1 & 87.2 & 90.6 \\
Dense ST adapter, dim=256 & 25M & 93.4 & 91.3 & 90.3 & 91.7 \\
D$^2$ST adapter \cite{pei2025d2st} & 9M & 92.6 & 88.6 & 87.6 &  89.6\\
\midrule
MoSTE & 19M & 95.5 & 90.7 & 91.3 & \textbf{92.5} \\
\bottomrule
\end{tabular}%
}
\end{table}

\subsection{Effect of Scaling}

To evaluate scalability, we follow recent large-scale training protocols based on DF40~\cite{yan2024df40}. Our model ($N=4$, Top-$k=3$) is trained on a DF40 subset containing over 25,000 video sequences spanning 23 distinct Face Swap and Reenactment methods. We ensure no overlap with other evaluation datasets. For comparison, we retrain EFFORT, a strong PEFT baseline, using the authors’ official implementation, increasing the residual rank to handle greater manipulation diversity. We also report results from DF40 project checkpoints for CLIP full fine-tuning \cite{radford2021learning} and an I3D-based \cite{carreira2017quo} video detector.

Evaluation is performed on four benchmarks, including FF++, DFD, CDF-v2, and FaceShifter. Table~\ref{tab:scaling} indicates that MoSTE benefits from large-scale and diverse training, while maintaining consistent generalization performance across unseen manipulations compared to SOTA.

\begin{table}[t]
\centering
\caption{Video-level AUC (\%) of different deepfake detection methods trained on DF40~\cite{yan2024df40}. $\dagger$ denotes models evaluated from the released DF40 checkpoints, while FR+FS indicates training on a combination of Face Swap and Face Reenactment sequences.}
\label{tab:scaling}
\setlength{\tabcolsep}{5pt}
\resizebox{\columnwidth}{!}{%
\begin{tabular}{l|c|c|c|c|c|c}
\toprule
Method & Train Set & FF++ & CDF-v2 & DFD & FSh & Avg \\
\midrule

CLIP$^\dagger$  & DF40 & 88.4 & \textbf{92.0} & 82.3 & 93.9 & 89.2 \\
EFFORT & FR+FS  & \textbf{95.7} & 87.0 & \underline{84.1} & 93.4 & \underline{90.1} \\
I3D $^\dagger$& FR+FS & \underline{95.2} & 76.3 & 68.0 & \underline{95.1} & 83.7 \\

Ours & FR+FS & 93.8 & \underline{91.5} & \textbf{96.7} & \textbf{95.7} & \textbf{94.4} \\
\bottomrule
\end{tabular}
}
\end{table}

\section{Conclusions}
In this paper, we introduced MoSTE, a novel Mixture of Spatio-Temporal Experts framework designed to improve generalization in video deepfake detection. Different manipulation types produce fundamentally different spatial and temporal artifacts, rendering a single adaptation pathway inherently biased toward particular forgery families, \textit{i.e.,} face swap and face reenactment.
To overcome this limitation, MoSTE decouples feature representation into parallel spatial and temporal experts.  Specifically, we learn a set of specialized experts and dynamically select which experts process each token. This enables the model to adapt to heterogeneous forgery patterns while remaining fully label-free and end-to-end trainable. 
In our extensive evaluations we showcase that our single-stage approach successfully balances detection capabilities across diverse manipulation methods. By establishing new state-of-the-art performance on both, cross-dataset and cross-manipulation benchmarks, alongside proven robustness to real-world video perturbations, MoSTE provides a highly adaptable detection approach in the continuously evolving landscape of video forgeries.

\section*{Acknowledgements}
This work was granted access to the HPC resources of IDRIS
 under the allocation 2026-AD011016744R1 and 2025-AD011015982R1 made by GENCI. The work of A.~Dantcheva has been supported by the French government, through the France 2030 plan managed by the National Research Agency (ANR) with the reference number "ANR-23-IACL-0001".
%
%

{
    \small
    \balance
    \bibliographystyle{ieeenat_fullname}
    \bibliography{main}
}

\end{document}